\documentclass[runningheads]{llncs}
\usepackage[T1]{fontenc}
\usepackage[utf8]{inputenc} 
\usepackage[acronym]{glossaries}
\usepackage{graphicx}
\usepackage{hyperref}
\usepackage{xcolor}
\usepackage{array}
\usepackage{booktabs}
\usepackage{tabularx}
\usepackage{float}
\usepackage{makecell}
\usepackage{subcaption}
\usepackage{enumitem}
\begin{document}

\newacronym{pbpm}{PBPM}{predictive business process monitoring}
\newacronym{llm}{LLM}{large language model}
\newacronym{bpm}{BPM}{business process management}
\newacronym{lstm}{LSTM}{long short-term memory}
\newacronym{cnn}{CNN}{convolutional neural network}
\newacronym{gnn}{GNN}{graph neural network}
\newacronym{mlp}{MLP}{multi-layer perceptron}
\newacronym{gan}{GAN}{generative adversarial network}
\newacronym{nlp}{NLP}{natural language processing}
\newacronym{rf}{RF}{random forest}
\newacronym{dt}{DT}{decision tree}
\newacronym{nb}{NB}{gaussian naïve bayes}
\newacronym{knn}{KNN}{k-nearest neighbor}
\newacronym{lr}{LR}{logistic regression}
\newacronym{svm}{SVM}{support vector machine}
\newacronym{xgb}{XGBoost}{extreme gradient boosting}
\newacronym{tr}{TR}{transformer}
\newacronym{ml}{ML}{machine learning}
\newcommand{\mypar}[1]{\vspace{0.5pt}\noindent\textbf{#1.}}
\newcommand{\mypartwo}[1]{\vspace{0.5pt}\noindent\textit{#1.}}

\newcommand{\etal}{\emph{~et~al.}}
\title{A Human-In-The-Loop Approach for Improving Fairness in Predictive Business Process Monitoring}
\titlerunning{Improving Fairness in Predictive Process Monitoring}
\author{
Martin K\"appel\inst{1}\orcidID{0009-0003-3420-8037}
\and 
Julian Neuberger\inst{2}\orcidID{0009-0008-4244-7659}
\and 
Felix M\"ohrlein\inst{2}\orcidID{0009-0004-7142-6102}
\and 
Sven Weinzierl\inst{1}\orcidID{0000-0003-2268-7352}
\and
Martin Matzner\inst{1}\orcidID{0000-0001-5244-3928}
\and
Stefan Jablonski\inst{2}
}
\authorrunning{K\"appel et al.}
% First names are abbreviated in the running head.
% If there are more than two authors, 'et al.' is used.
%
\institute{%
    Friedrich-Alexander-University Erlangen-Nuremberg, Germany
    \email{firstname.lastname@fau.de}
\and
    University of Bayreuth, Bayreuth, Germany 
    \email{firstname.lastname@uni-bayreuth.de}
}
\maketitle              % typeset the header of the contribution
\begin{abstract}
Predictive process monitoring enables organizations to proactively react and intervene in running instances of a business process.
Given an incomplete process instance, predictions about the outcome, next activity, or remaining time are created.
This is done by powerful machine learning models, which have shown impressive predictive performance.
However, the data-driven nature of these models makes them susceptible to finding unfair, biased, or unethical patterns in the data. Such patterns lead to biased predictions based on so-called sensitive attributes, such as the gender or age of process participants. Previous work has identified this problem and offered solutions that mitigate biases by removing sensitive attributes entirely from the process instance. However, sensitive attributes can be used both fairly and unfairly in the same process instance. For example, during a medical process, treatment decisions could be based on gender, while the decision to accept a patient should not be based on gender. This paper proposes a novel, model-agnostic approach for identifying and rectifying biased decisions in predictive business process monitoring models, even when the same sensitive attribute is used both fairly and unfairly. The proposed approach uses a human-in-the-loop approach to differentiate between fair and unfair decisions through simple alterations on a decision tree model distilled from the original prediction model. Our results show that the proposed approach achieves a promising tradeoff between fairness and accuracy in the presence of biased data. All source code and data are publicly available at \url{https://doi.org/10.5281/zenodo.15387576}.

\end{abstract}
%
%%%%%%%%%%%%%%%%%%%%%%%%%%%%%%%%%%%%%%%%%%%%%
% INTRODUCTION
%%%%%%%%%%%%%%%%%%%%%%%%%%%%%%%%%%%%%%%%%%%%%
\vspace{-0.7cm}
\section{Introduction}
\label{sec:introduction}

\Gls{pbpm}~\cite{maggi2014predictive,grigori2004business} has been established as an important subfield in business process management research. \gls{pbpm} supports process stakeholders at run-time by predicting how a process instance will evolve up to their completion~\cite{di2022predictive}. 
Predictive models for this task are often built from historical event data using \gls{ml} algorithms \cite{weinzierl2024machine} and predict properties related to, among others, behavior (e.g., next activities~\cite{tama2019empirical}), outcome (e.g., expected outcome and performance~\cite{teinemaa2018outcomeoriented}), or time (e.g., remaining time~\cite{verenich2019survey} or delays). 
From a business perspective, predictions of these properties offer considerable added-value potential through proactive intervention in running instances of a business process to mitigate risks or avoid undesired process outcomes \cite{marquez2017predictive}. As prediction models can be used in business processes, in which critical decisions need to be made, such as the transfer of a patient to intensive care in a patient treatment process~\cite{Zilker2024}, questions about their fairness and the ethical implications of their use have come to the forefront~\cite{de2024achieving}. Fairness concerns the equitable treatment of individuals or groups~\cite{dwork2011fairnessawareness}, particularly in the presence of sensitive attributes, such as gender, age, race, or socioeconomic status~\cite{Oneto2020}. 
These attributes often correlate with historical inequalities or systemic biases embedded in the data~\cite{Barocas2016}. 
However, \gls{ml} models are designed to optimize predictive performance by learning patterns from data, without regard to fairness.
Consequently, models learning unfair patterns can improve predictive performance, while also risking perpetuating or even amplifying existing inequities.

\begin{figure}[tb]
    \centering
    \includegraphics[width=0.85\linewidth]{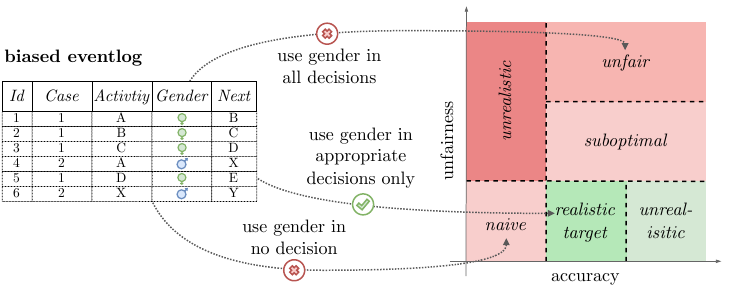}
    \caption{\small Using sensitive attributes increases the accuracy of predictions on biased data, but potentially decreases fairness. Ignoring sensitive attributes will guarantee fairness, but decrease predictive performance more than necessary. Our approach finds a balance between the two extremes. Note that some of the combinations are unrealistic, i.e., a prediction model without access to sensitive attributes will not be unfair, and a fair prediction model will not outperform an unfair prediction model, when prediction accuracy is evaluated on an unfairly biased event log.}
    \label{fig:motivation}
    \vspace{-0.6cm}
\end{figure}

Fairness in AI is violated by so-called biases, which are defined as systematic deviations of estimated parameters from true values \cite{feuerriegel2020fair}. 
As these deviations can be generally positive or negative, some biases can lead to discrimination, and others cannot be inherently harmful, as they may be necessary for the model to achieve its intended purpose. 
Fig.~\ref{fig:motivation} illustrates different scenarios when dealing with sensitive attributes in \gls{pbpm}.
For instance, in the domain of healthcare, it might be justified to use a patient's gender for determining effective treatments and drug prescriptions, but discriminating when deciding if their request for treatment should be approved.
Consequently, removing the influence of the sensitive attributes entirely from the model is an inadequate solution and would lead to suboptimal predictions.

This problem is aggravated by the fact that most of the used advanced machine learning models (e.g., deep learning models) largely represent a black box, i.e., internal prediction making is intransparent~\cite{Nauta2023}. 
This lack of interpretability prevents stakeholders and domain experts from verifying the ethical use of sensitive attributes in a given prediction model.
Hence, this paper explores the following research question: 
\textit{How can discriminatory biases be identified and systematically mitigated in \gls{pbpm} models to ensure fair decision-making, even when the same sensitive attribute is used both fairly and unfairly?}
By answering this central research question, we make the following three core contributions.

\begin{enumerate}[topsep=0pt,itemindent=0pt,label=\textbf{C\arabic*}]
    \item 
    We explore how the same sensitive attribute can lead to both fairly and unfairly biased decisions in the same process instance.

    \item 
    We enable domain experts to identify unfair use of sensitive attributes in prediction models and revise them, without unnecessarily sacrificing predictive performance.

    \item 
    We provide three datasets with novel characteristics to facilitate the development of new approaches for improving fairness in \gls{pbpm}.
\end{enumerate}

\noindent
The remainder of the paper is structured as follows.
Sect.~\ref{sec:preliminaries} defines basic terminology.
In Sect.~\ref{sec:related-work}, we discuss how previous work relates to our approach.
We then introduce our proposed method for debiasing prediction models in Sect.~\ref{sec:approach}. 
In Sect.~\ref{sec:experiment-setup}, we evaluate our approach from different perspectives. 
Sect.~\ref{sec:conclusion} concludes the paper by discussing implications, limitations, and future work.
%
%
%%%%%%%%%%%%%%%%%%%%%%%%%%%%%%%%%%%%%%%%%%%%%
% PRELIMINARIES
%%%%%%%%%%%%%%%%%%%%%%%%%%%%%%%%%%%%%%%%%%%%%

\section{Preliminaries}
\label{sec:preliminaries}
The main input of a \gls{pbpm} approach is an \emph{event log}, i.e., a set of records of the execution of a business process~\cite{vanderAalst.2016}. 
A \emph{process instance} represents an execution of the business process and contains events that record the execution of an activity, i.e., a well-defined step in a business process~\cite{vanderAalst.2016}. Each event can be formally described as follows:
\begin{definition}
An \textbf{event} is a tuple $e = (c, a, t, d)$, where $c$ is a case identifier linking the event to a particular process instance, $a$ is the name of the activity that has been executed, $t$ is the timestamp denoting when the event occurred, and $d$ are additional event attributes, collectively referred to as the events' data payload.    
\end{definition}
For each event attribute, we define a mapping $\pi_{p}$ that returns the value of an event attribute $p$ for a given event. If an event $e$ does not record or does not have a value for an attribute $p$, $\pi_p(e)$ returns $\bot$. All events related to the same process instance, ordered by their timestamp, form a so-called \emph{trace} (also known as a \emph{case})~\cite{vanderAalst.2016}. 
\begin{definition}
A \textbf{trace} $\sigma = \langle e_1, \dots, e_n \rangle$ is a finite sequence of events such that for $1 \leq i < j \leq n$ holds that all events are ordered according to their timestamp (i.e., $\pi_t(e_j) \geq \pi_t(e_i)$) and all events belong to the same process instance (i.e., $\pi_c(e_j) = \pi_c(e_i)$). 
\end{definition}
Traces, similar to events, possess attributes, which are referred to as case attributes to distinguish them. For a trace $\sigma$ and a case attribute $r$, we define a mapping $\mu_r$ that returns the value of the particular case attribute for $\sigma$. If a trace $\sigma$ does not record a case attribute $r$, $\mu_r(\sigma)$ returns $\bot$.

An \emph{event log} $L$ is a collection of traces. To represent the instances of running processes at different points in time, we utilize the prefixes of a trace~\cite{vanderAalst.2016}: 
\begin{definition}
Let $\sigma = \langle e_1, \dots, e_n \rangle$ be a trace. The \textbf{prefix} of trace $\sigma$ of length $l \in \{1,\dots,n-1\}$ is defined as a function $hd$ that returns the first $l$ events of $\sigma$, i.e., $hd(\sigma, l) = \langle e_1, \dots, e_l\rangle$.     
\end{definition}

Next activity prediction is defined as a function $\Omega$ that, given a prefix $\langle e_1, \dots, e_l\rangle$, predicts the activity of the next event $e_{l+1}$, which is not yet known. Most of the \gls{pbpm} techniques employ \gls{ml} algorithms to train a \emph{prediction model} to learn the function $\Omega$ based on the prefixes extracted from an event log $L$.
Depending on whether the used prediction model provides direct insights into the reasoning or not, we distinguish between \emph{white box} and \emph{black box} models.

%%%%%%%%%%%%%%%%%%%%%%%%%%%%%%%%%%%%%%%%%%%%%
% RELATED WORK
%%%%%%%%%%%%%%%%%%%%%%%%%%%%%%%%%%%%%%%%%%%%%

\section{Related Work}
\label{sec:related-work}
Although the field of fairness in \gls{ml} has led to the emergence of a plethora of approaches and methods as witnessed in the surveys of \emph{Caton}\etal~\cite{Caton2024} and \emph{Pessach}\etal~\cite{Pessach2022}, in the context of business process management, and especially in \gls{pbpm}, this research stream is still in its infancy. Apart from emphasizing the general importance of fairness (or in a broader sense ethics) in process management (e.g.,~\cite{Kern2024,deArteaga2022}), there is still little research on this topic. In \emph{Pohl}\etal~\cite{Pohl2023}, existing fairness concepts in \gls{ml} are categorized and their relevance to process mining research is assessed. Other process mining works focus, for example, on developing a process discovery algorithm optimized for group fairness~\cite{Muskan2025} or enriching process models by incorporating fairness considerations~\cite{Amico2025}. 
However, in \gls{pbpm}, where methods are often considered process mining approaches, three works tackle the topic of fairness. We review these works in the following.

\emph{Qafari}\etal~\cite{qafari2019fairness} represent one of the initial efforts to incorporate fairness into \gls{pbpm}. 
The authors utilize discrimination-aware decision trees, which extend traditional decision trees by introducing fairness constraints. 
Specifically, when the model’s predictions exceed a certain threshold of unfairness, i.e., when they are disproportionately influenced by sensitive attributes, a relabeling technique is applied. 
This method of relabeling adjusts model predictions, such that the discriminatory impact of sensitive attributes is reduced while keeping predictive performance as high as possible. 
Although the aim of this technique is similar to our approach, there is a key difference: 
Relabeling in~\cite{qafari2019fairness} is performed automatically, based on the assumption that all biases are undesirable and should be eliminated regardless of the context. 
In contrast, our approach emphasizes the involvement of domain experts to decide which biases are unwanted, allowing for the preservation of nuanced, context-dependent correlations.

In \emph{de Leonie}\etal~\cite{de2024achieving}, biases in predictive models are being addressed by leveraging adversarial learning~\cite{goodfellow2014}, a method where two \gls{ml} models are trained simultaneously: a predictor and an adversary model. The predictor model focuses on producing accurate predictions, while the adversary model tries to identify sensitive attributes based on the output of the predictor. The predictor is penalized when the adversary succeeds, encouraging fairness by reducing reliance on these attributes. While effective at reducing biases, this penalty-based approach lacks context sensitivity, treating all biases as inherently undesirable. This indiscriminate removal of biases may unintentionally eliminate meaningful correlations, thereby diminishing the predictive model’s utility in specific scenarios where such relationships are crucial.

Last, \emph{Peeperkorn}\etal~\cite{peeperkorn2024} propose ensuring independence between predictions and sensitive group memberships by using a composite loss function for training the \gls{ml} model. 
Compared to traditional loss functions, which seek to only optimize the predictive performance of an \gls{ml} model, this composite loss function incorporates integral probability metrics (e.g., Wasserstein distance)~\cite{Panaretos2019} to find a balance between accuracy and fairness. 
However, similar to the other approaches, the uniform elimination of biases may inadvertently remove desirable correlations, potentially affecting the utility of the predictive model in certain contexts.

In contrast to these three fairness-oriented \gls{pbpm} approaches, our proposed approach i) does not generally consider biases in the data as inherently negative and ii) incorporates a human expert to review changes before automatically applying them. 

%
%%%%%%%%%%%%%%%%%%%%%%%%%%%%%%%%%%%%%%%%%%%%%
% APPROACH
%%%%%%%%%%%%%%%%%%%%%%%%%%%%%%%%%%%%%%%%%%%%%

\section{Approach}
\label{sec:approach}

\begin{figure}[bt]
    \centering
    \includegraphics[width=0.95\linewidth]{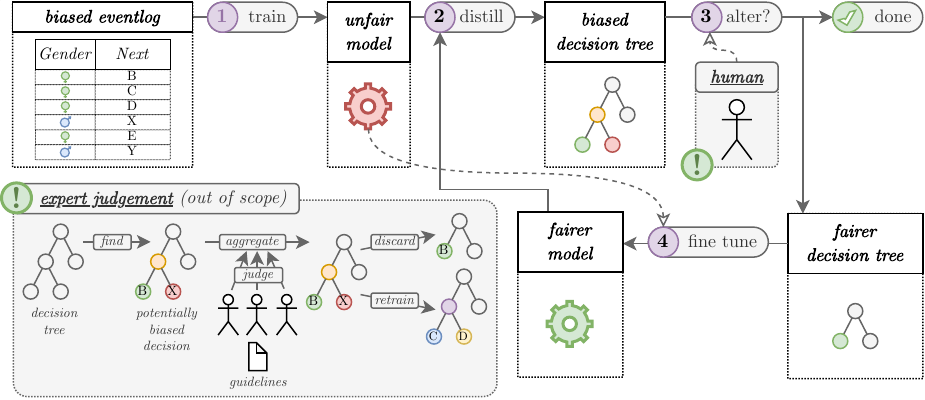}
    \caption{\small Overview of the proposed approach.}
    \label{fig:approach-overview}

\end{figure}

Our approach consists of four steps, which we outline and visualize in Fig.~\ref{fig:approach-overview}.

\subsection{Training}
\label{sec:approach:training}
First, a prediction model is trained using a (potentially) biased event log, containing process instances with unfairly biased decisions based on sensitive attributes.
We treat the prediction model as a black box, i.e., it should be trainable given an event log in a common format, such as XES\footnote{\url{http://xes-standard.org/ieee2016}, last accessed Feb. 26, 2025.}, and should produce predictions given an incomplete trace (prefix).
We further assume that this model is able to use case-level attributes for predictions.
See Sect.~\ref{sec:experiment-setup:model} for a description of the prediction model we used in our experiments.

To use such a prediction model in our approach, it is trained without any adjustments, i.e., following the recommended training regime.
This first step thus results in a prediction model $M$ that approximates the function $\Omega$ defined in Sect.~\ref{sec:preliminaries}.

\subsection{Distillation}
\label{sec:approach:distillation}
Knowledge distillation transfers knowledge from a large, complex model, called \textit{teacher} model, into a smaller, more efficient \textit{student} model~\cite{hinton2015distilling}.
Instead of using the targets given in the original training data (event log), the student model is instead trained using the outputs of the teacher model.
Compared to their respective teacher models, student models are able to achieve similar predictive performance with far less computational complexity, i.e., parameters, since they merely have to learn to imitate teacher behavior, instead of learning rules from data directly.
In our approach, we use knowledge distillation as a means of making large \gls{pbpm} black-box models \textit{interpretable}.
Training a black-box model first and then distilling its knowledge into a white-box model has been shown to outperform direct training of the white-box model on the same data~\cite{liu2018improving}.

We therefore distill the knowledge contained in the prediction model $M$ (teacher) into a decision tree model $D$ (student), which serves as a means of explaining the decision rules the prediction model learned. 
Decision tree models are directly interpretable, white-box \gls{ml} models, well suited for \textit{local} interpretation, i.e., understanding how a model comes to a prediction for a given input~\cite{slack2019assessing}, and which features in the data contribute to the prediction.
In our approach, the model $D$ thus serves as a common interface between the complex, black-box prediction model $M$ and a human process expert who wishes to evaluate learned prediction rules in terms of their fairness.

The model $D$ still contains unfairly biased decisions learned from the event log, which we remove in the next step of our approach.

\subsection{Alteration}
\label{sec:approach:alteration}
\begin{figure}[bt]
    \centering
    \includegraphics[width=0.8\linewidth]{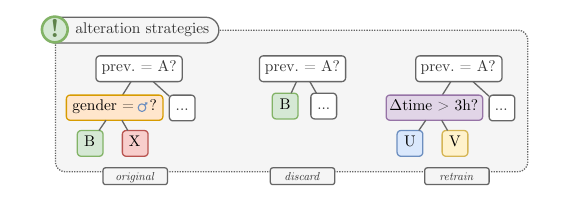}
    \caption{\small The two strategies used to alter the distilled decision tree model $D$ in our approach.}
    \label{fig:alteration-strategies}
    \vspace{-0.3cm}
\end{figure}

We look for inner nodes that use sensitive attributes as the basis for their splits and assess whether such usage is justified in the context of the node.
Human experts review such potentially unfair biasing decision nodes in model $D$, possibly using internal company guidelines and best practices, if they are documented.
If multiple experts are involved, their individual judgments are aggregated into a final one\footnote{Note, that -- for reasons of focus and space in the paper -- our experiments assume agreement between experts was already found, and all unfairly biasing decision nodes have been identified.
We plan to research how to efficiently find consensus in human experts through proper tool support in future work.}.
If no node representing a potentially unfair decision is found, or the human experts decide against removing it, our approach terminates here, and model $M$ remains unchanged.
Otherwise, the experts can delete any nodes they determine to unfairly rely on sensitive attributes, resulting in a fairer decision tree model $D^*$.
To this end, we propose making alterations following one of two strategies, illustrated in Fig.~\ref{fig:alteration-strategies}.

First, if a node in the decision tree model is deleted, one of the two sub-trees is selected to replace it, as shown in Fig.~\ref{fig:alteration-strategies} alteration strategy \textit{discard}, where the node is removed and replaced with the left sub-tree \textit{B}.
This strategy is guaranteed to remove an unfair decision based on a given sensitive attribute, while preserving decisions based on the same attribute in the sub-tree.
Yet, it potentially alters predicted process behavior significantly, as the discarded sub-tree might still contain valid decisions.

Therefore, if the offending node and its corresponding sub-tree are deleted, we can reconstruct the sub-tree without using the sensitive attribute, as shown in Fig.~\ref{fig:alteration-strategies} in the \textit{retrain} strategy.
When multiple sensitive attributes are present in the event log, the retraining strategy may result in new, unfairly biased decisions using the sensitive attributes remaining after node deletion.
Therefore, decisions in model $D^*$ are reviewed again before proceeding to the next step in our approach.

Simply removing sensitive attributes from decisions can shift the bias to non-sensitive attributes acting as \emph{proxy attributes} instead of removing it~\cite{feuerriegel2020fair,panda2022don}.
In our approach, this effect would be obvious to the human expert in the next iteration, where they could remove such proxy attributes as well.
After the \textit{Alteration} step, we are left with a revised, \textit{fairer} decision tree model $D^*$, that is used for fine-tuning the original \gls{pbpm} model $M$ in the next step.

\subsection{Fine Tuning}
\label{sec:approach:fine-tuning}
We then use model $D^*$ to fine tune model $M$, i.e., adapt the trained model $M$ with new (revised) data, to retain useful features and decisions.
We fine tune the model in an effort to only update unfair decision rules learned by the model, without adjusting anything else, preserving its predictive performance.

Accordingly, instead of using the training targets given in the original event log, we let model $D^*$ predict the next activity and use this prediction as a training target for model $M$.
Since a distilled model is inherently less expressive than the original model, using the predictions of model $D^*$ for \textit{all} prefixes could degrade the predictive performance of model $M$ in cases where no fairness concerns are present.
Alternatively, we can use revised predictions only for prefixes where the predictions of the models $D$ and $D^*$ differ, i.e., were affected by expert alterations in the previous step.
Our approach tries both methods and picks the one that degrades the prediction performance of model $M$ less, while reducing unfairness.

After fine tuning the model $M$, we are left with a fairer prediction model $M^*$. 
This model can be distilled again for the human expert to verify that their alterations have had the expected effect, and check if additional fairness improvements can be made.
%
%
%
%%%%%%%%%%%%%%%%%%%%%%%%%%%%%%%%%%%%%%%%%%%%%
% EXPERIMENT SETUP
%%%%%%%%%%%%%%%%%%%%%%%%%%%%%%%%%%%%%%%%%%%%%

\section{Experiment Setup}
\label{sec:experiment-setup}
Due to the fact that other existing approaches completely remove sensitive attributes, we cannot directly compare our approach with them as the objective is fundamentally different (see Sect.~\ref{sec:related-work}). 
However, we use the underlying idea of these approaches for the comparison and compare ourselves with a fair baseline model $F$, in which all sensitive attributes have been removed. As this paper assumes that agreement on the bias between experts was already found, our evaluation focuses on the feasibility of removing the bias.

\subsection{Evaluation Data}
\label{sec:experiment-setup:data}
To evaluate the effects of our approach, we use three different event logs that contain sensitive attributes and use them unfairly.
Real-life event logs containing discriminating process instances of an organization are not publicly available, as their very nature discourages publication. 
Additionally, to properly evaluate the benefits of our method, we require sensitive attributes causing both acceptable and unacceptable biases in a process instance.
For this purpose, we generate a new \textit{synthetic} event log and use two existing, well-studied real-life event logs, \textit{enriched} with sensitive case-level attributes.

\mypar{Simulated Cancer Screening Process (\textit{cs})}
This synthetic event log is the result of simulating a process of moderate complexity, containing three decisions where the sensitive attributes \textit{age} and \textit{gender} influence the outcome of a decision.
Except for one decision, none of these decisions are medically justified, and thus introduce negative bias. We simulated $10,000$ cases, sampled the value for the \textit{age} attribute from a normal distribution with $\mu=45$ and $\sigma=10$, and also constrained the attribute values to the range $[20;85]$.
Further, for this example, values for the \textit{gender} attribute are uniformly sampled from the attributes \textit{male} and \textit{female}.
These attributes are then used within the decisions. For example, $gender=female$ will always result in mammary cancer screening\footnote{Note: Even though men can get breast cancer, in Germany at least, routine screening for breast cancer is only carried out for women.}, or $age=50$ will result in informing the patient of prevention measures in $70\%$ of cases and explaining the diagnosis and discussing treatment options in the remaining $30\%$.

Furthermore, to enrich existing real-life event logs, we add sensitive attributes in such a way that they enable explaining certain workflow decisions during a process instance.
Note that these decisions and the resulting traces already exist in the event logs. Thus, we only choose attribute values that correlate with the recorded behavior.

\mypar{BPI Challenge 2012 (\textit{bpi})}
This event log comprises 13,087 cases, sourced from a Dutch financial institution. 
It contains multiple subprocesses, of which we use the \textit{A\_} subprocess describing loan application procedures.\footnote{\url{https://data.4tu.nl/articles/dataset/BPI\_Challenge\_2012/12689204}, last accessed Feb. 13, 2025.}
To explore fairness concerns, we enrich the event log with the categorical sensitive case attribute \textit{gender}, such that it (partially) explains two decisions in the loan application process.
Additionally, we use the case attribute \textit{amount required}, which denotes the size of the requested loan.

\begin{enumerate}
    \item 
    After a loan request was \textit{partly submitted} (A\_PARTLY\_SUBMITTED), it is \textit{pre-accepted} (event A\_PREACCPETED) in $56.3\%$ cases and \textit{declined} (event A\_DECLINED) in the remaining $43.7\%$.
    We assign cases, that were pre-accepted, $gender=female$ with a probability of $70\%$, and cases that were declined, $gender=male$ with a probability of $70\%$.
    Note that this assignment is arbitrary and only serves as a way of making the decision between pre-accepting and declining a loan request explainable.
    For the purpose of our experiments, we consider this decision \underline{unfairly biased}.
    \item 
    If the request was ultimately \textit{approved} (event A\_APPROVED), we assign $gender=female$ with a probability of $70\%$, and $gender=male$ with a probability of $30\%$, whereas if it was \textit{canceled} (event A\_CANCELLED), this assignment was reversed.
    Note that this assignment is again arbitrary and only serves as a way of making the decision explainable.
    We \underline{do not} consider this decision as unfairly biased.
\end{enumerate}

\noindent
All remaining cases are assigned $gender=female$ and $gender=male$ with equal probability.
This ruleset results in a total of $56.6\%$ male and $43.4\%$ female applicants.

\mypar{Hospital Billing (\textit{hb})}
Following Qafari and van der Aalst~\cite{qafari2019fairness}, we select the last 20,000 traces from the Hospital Billing (\textbf{\textit{hb}}) event log~\cite{mannhardt2017hospital}.
To these traces, we add the case-level attributes \textit{age} (continuous) and \textit{gender} (categorical).
Controlling when these attributes create unfair biases, let us design four scenarios:

\begin{enumerate}
    \item \textbf{$-$age, $-$gender}: no unfair bias in age and gender,
    \item \textbf{$-$age, +gender}: no unfair bias in age, unfair bias in gender,
    \item \textbf{+age, $-$gender}: unfair bias in age, no unfair bias in gender,
    \item \textbf{+age, +gender}: unfair bias in both age and gender.
\end{enumerate}

These four scenarios help us to investigate the degree to which our approach can identify unfairly biased decisions in the presence of additional, non-biasing case-level attributes.
In addition, the scenario \textbf{$-$age, $-$gender} allows us to demonstrate that our approach has no detrimental effect on prediction performance, when there are no negative biases in the data. 
\subsection{Metrics}
\label{sec:experiment-setup:metrics}
To measure the quality of predictions made by the prediction model, we use \textit{accuracy}, which is a commonly used metric for measuring the predictive performance of models in \gls{pbpm} \cite{marquez2017predictive}.
It is defined as the number of predictions by the model that match the expected next activity in the event log.
For measuring the fairness of these predictions, we use the \textit{demographic parity} (DP) metric, which is a common metric in the literature on fairness in \gls{ml} \cite{castelnovo2022clarification,barocas2023fairness}.
This metric assumes that a sensitive attribute $S$ can be split into two demographic groups $s_1$ and $s_2$. 
The probability for a desired positive prediction $y=1$, given the demographic group $s_i$, is then $P_i=P(y=1|S=s_i)$.
A model satisfies demographic parity, if $P_1=P_2$, i.e., both demographic groups are equally likely to receive positive predictions.
Conversely, the difference $\Delta{DP}=\left|P_1-P_2\right|$ between the two probabilities measures \textit{unfairness}, and should be minimized.
Since the true probabilities are unknown, they are estimated from the model predictions on the event log, i.e., as the ratio between positive predictions for a demographic group compared to the total number of predictions for that group.

\subsection{Prediction Model}
\label{sec:experiment-setup:model}
We use a feedforward neural network model, implemented in Python and Keras\footnote{See \url{https://keras.io/}, last accessed Feb. 13, 2025.}, as a prediction model for all three event logs presented in Sect.~\ref{sec:experiment-setup:data}.
The model contains four hidden, fully connected layers of 512, 256, 128, and 64 neurons, each using the \textit{ReLU} activation function.
The output layer has one neuron for each unique activity that appears in the event log and uses a softmax activation function.

To obtain training examples, we slide a window of length 3 over all traces in the event logs, and pad traces that are shorter than 3.
All windows are then encoded in the following manner: 
Categorical attributes, including the \textit{activities} appearing in the window, are one-hot encoded, e.g., given an activity with index $i$ in the list of all $n$ activities appearing in the log, we encode it as a vector of length $n$, where all entries are $0$, except entry $i$, which is $1$. 
All numerical attributes, including the time elapsed since the last event in the window, are normalized to the range $[0,1]$. 
This prevents attributes with large magnitudes from dominating attributes with smaller magnitudes.
We then concatenate the encoding results to a single vector $\mathbf{x}$, which serves as the input to our neural network.
The corresponding training target $\mathbf{y}$ is the one-hot encoded next activity to the right of the window encoded in $\mathbf{x}$.

We train the model for a total of 10 epochs with a batch size of 32, using the \textit{categorical cross-entropy} loss function, and the \textit{Adam} optimizer with a learning rate of $0.001$. 
In our experiments, 10 epochs were sufficient for the model to converge without overfitting.
%
%
%
%%%%%%%%%%%%%%%%%%%%%%%%%%%%%%%%%%%%%%%%%%%%%
% RESULTS
%%%%%%%%%%%%%%%%%%%%%%%%%%%%%%%%%%%%%%%%%%%%%

\section{Results}
\label{sec:results}
\begin{figure}[bt]
\centering
\begin{subfigure}[t]{.50\textwidth}
    \centering
    \includegraphics[width=\linewidth]{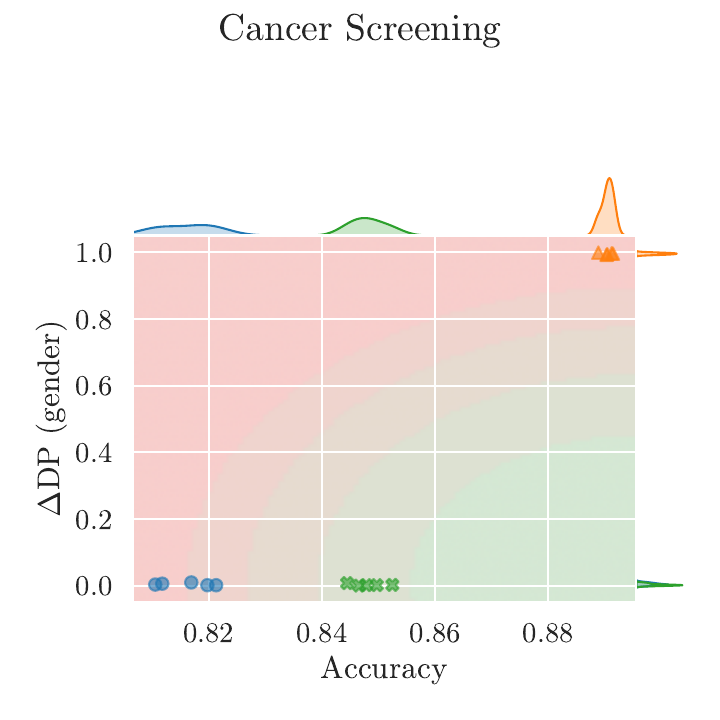}
    \caption{}
    \label{fig:results-cs}
\end{subfigure}%
\hfill%
\begin{subfigure}[t]{.50\textwidth}
    \centering
    \includegraphics[width=\linewidth]{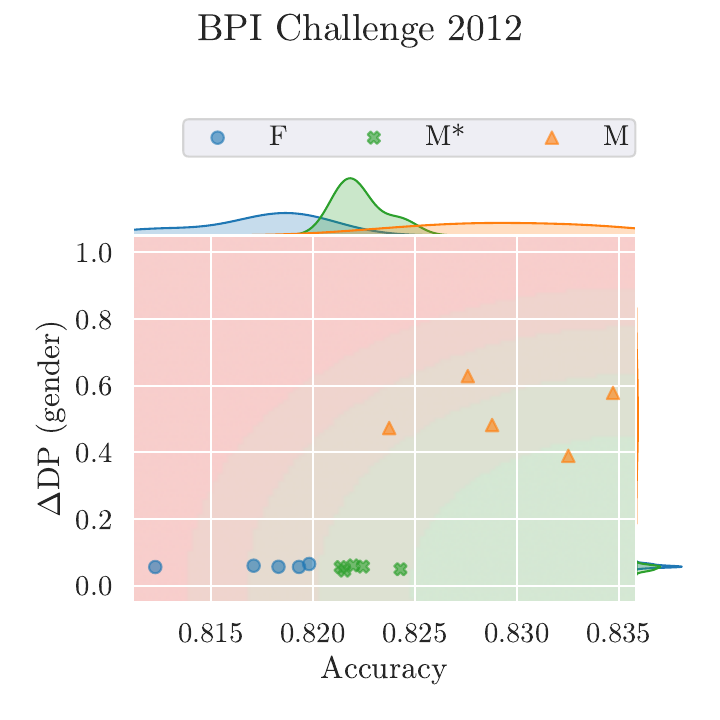}
    \caption{}
    \label{fig:results-bpi}
\end{subfigure}%
\caption{\small Tradeoff between accuracy and fairness on \textit{(a)} simulated Cancer Screening event log, and \textit{(b)} BPI Challenge 2012 event log, enriched with sensitive attributes introducing bias. 
Density estimates show how observations of accuracy (top) and unfairness (right) are distributed for the three models.
Even though the effect is weaker on the real-world BPI event log, our approach $M^*$ (green) reduces unfairness to minimal levels, while improving accuracy compared to model $F$ (blue), not using any sensitive attributes.
Model $M$ (orange) achieves higher accuracy than our approach, only by unfairly biasing decisions, i.e., high $\Delta{DP}$ values.}
\label{fig:results-cs-and-bpi}

\end{figure}

\mypar{Cancer Screening (Synthetic Data)}
Due to the simple nature of the \textit{cs} event log's underlying process model
and the strong influence of sensitive attributes on process transitions, this dataset produces the clearest results, as reported in Table \ref{tab:evaluation_results}.
Figure \ref{fig:results-cs} shows that the modified prediction model $M^*$ successfully achieves a notable improvement in accuracy over the baseline model $F$,
while effectively reducing $\Delta \textit{DP}$ to almost 0.
This $\Delta \textit{DP}$ of near 0 implies completely fair decisions, meaning no discrimination occurs based on the attribute \textit{gender}.
Notably, the modified prediction model $M^*$ exhibits worse accuracy than the original prediction model $M$.
However, this decrease in accuracy is to be expected, as we trained the model on an event log that was purposefully biased.
By enforcing fairness, model $M^*$ actively disregards these biased patterns, leading to decisions that contradict the original data distribution.
\begin{figure}[bt]
\centering
\begin{subfigure}[t]{.50\textwidth}
    \centering
    \includegraphics[width=\linewidth]{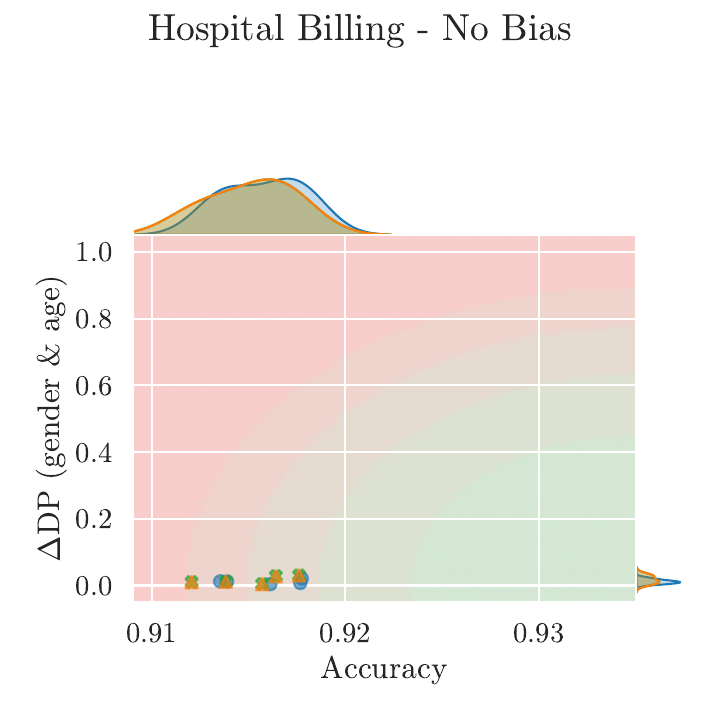}
    \caption{}
    \label{fig:results-hb-no-bias}
\end{subfigure}%
\hfill%
\begin{subfigure}[t]{.50\textwidth}
    \centering
    \includegraphics[width=\linewidth]{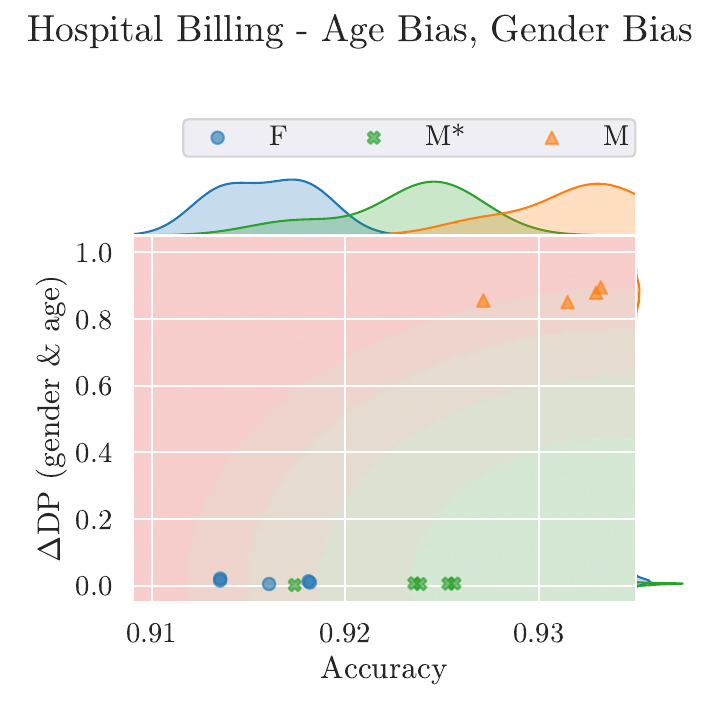}
    \caption{}
    \label{fig:results-hb-both-bias}
\end{subfigure}%
\caption{\small Tradeoff between accuracy and fairness on Hospital Billing event log, enriched with sensitive attributes that introduce \textit{(a)} \textbf{no} bias, and \textit{(b)} biases in both \textbf{age and gender}.}

\label{fig:results-hb-no-bias-and-both-bias}
\end{figure}

\mypar{BPI Challenge 2012}
In the \textit{bpi} event log, our approach also successfully reduces $\Delta \textit{DP}$ to baseline level, while maintaining a higher accuracy than model $F$, as presented in Fig.~\ref{fig:results-bpi}.
Notably, however, as reported in Table \ref{tab:evaluation_results}, the accuracy differences between models are much less pronounced in contrast to the \textit{cs} event log, causing the accuracy distributions to overlap.
Since accuracy is computed over a larger number of activities and the attribute \textit{gender} only influences a few transitions, its overall impact is relatively weaker.
Additionally, the attribute \textit{AMOUNT\_REQ} is available to all models and explains much of the variation in the transitions.
Meanwhile, the $\Delta \textit{DP}$ baseline is no longer as close to 0 as for the \textit{cs} event log.
This can be attributed to higher amounts of noise present in a real-life event log compared to our synthetic one.
\begin{table}[t]
    %\footnotesize
    \scriptsize
    \centering
    \begin{tabular}{c | ccc | ccc}
        \toprule
        \textbf{Event Log} & \multicolumn{3}{c|}{\textbf{Accuracy}} & \multicolumn{3}{c}{\textbf{$\Delta \textit{DP}$}} \\
        & $F$ & $M^*$ & $M$ & $F$ & $M^*$ & $M$ \\
        \midrule
        cs &  .816 $\pm$ .005 &  .848 $\pm$ .003 &  .890 $\pm$ .001 &  .004 $\pm$ .004 &  .002 $\pm$ .003 &  .996 $\pm$ .002 \\
        hb ($-$age $-$gender) &  .916 $\pm$ .002 &  .915 $\pm$ .002 &  .915 $\pm$ .002 &  .011 $\pm$ .004 &  .017 $\pm$ .005 &  .017 $\pm$ .005 \\
        hb ($-$age +gender) &  .916 $\pm$ .002 &  .920 $\pm$ .004 &  .926 $\pm$ .003 &  .009 $\pm$ .001 &  .036 $\pm$ .039 &  .439 $\pm$ .563 \\
        hb (+age $-$gender) &  .916 $\pm$ .002 &  .920 $\pm$ .003 &  .928 $\pm$ .004 &  .026 $\pm$ .016 &  .032 $\pm$ .002 &  .478 $\pm$ .602 \\
        hb (+age +gender) &  .916 $\pm$ .002 &  .923 $\pm$ .003 &  .932 $\pm$ .003 &  .012 $\pm$ .008 &  .005 $\pm$ .001 &  .877 $\pm$ .057 \\
        bpi &  .817 $\pm$ .003 &  .822 $\pm$ .001 &  .829 $\pm$ .004 &  .058 $\pm$ .004 &  .053 $\pm$ .006 &  .510 $\pm$ .094 \\
        \bottomrule
    \end{tabular}
    \caption{\small Evaluation of accuracy and $\Delta \textit{DP}$ for the \textit{Cancer Screening (cs)} log,
    the \textit{BPI Challenge 2012 (bpi)} log, and four versions of the \textit{Hospital Billing (hb)} log,
    where the attributes \textit{age} and \textit{gender} are annotated based on whether they introduce a bias ($+$) or not ($-$).
    Since the \textit{hb} event log uses two sensitive attributes, we report their average $\Delta \textit{DP}$.
    The reported values represent the mean ($\mu$) and standard deviation ($\sigma$) across five validation folds, expressed as $\mu \pm \sigma$.
    }
  
    \label{tab:evaluation_results}
\end{table}

%\vspace{-0.2cm}
\mypar{Hospital Billing}
Since the majority of traces in the \textit{hb} event log follow the same path,
the next activity is highly predictable even without access to the case attributes.
Therefore, similar to the \textit{bpi} dataset, the observed accuracy differences between models in the \textit{hb} event log are minimal, as reported in Table \ref{tab:evaluation_results}.
If \textbf{neither attribute} is biased, the model $M^*$ performs identically to the models $M$ and $F$.
Since the distilled decision tree model contains no inner nodes that split based on sensitive attributes, no modifications take place.
This demonstrates that our approach does not unnecessarily reduce accuracy when no unfairness is present.
When only \textbf{one attribute} is biased, as visualized in Fig.~\ref{fig:results-hb-no-bias}, results align with those observed in the \textit{cs} and \textit{bpi} event logs.
The model $M^*$ reduces $\Delta \textit{DP}$ for the biased attribute close to baseline levels while maintaining better accuracy. The inclusion of the unbiased attribute does not introduce unfairness, considering that its $\Delta \textit{DP}$ is low for all models.
However, occasional outliers are observed where $\Delta \textit{DP}$ for the biased attribute is not reduced as much as expected. 
In practice, such cases likely require adjustments in hyperparameters during fine tuning, such as increasing the number of epochs or the learning rate.
If \textbf{both attributes} introduce bias, their combined effect amplifies the disparity in accuracy between the models, with the accuracy of model $M^*$ still in between that of models $F$ and $M$,
as presented in Fig.~\ref{fig:results-hb-both-bias}.
The $\Delta \textit{DP}$ of model $M^*$ is still significantly reduced for both attributes compared to model $M$.

\begin{figure}[bt]
    \centering
    \includegraphics[width=1\linewidth]{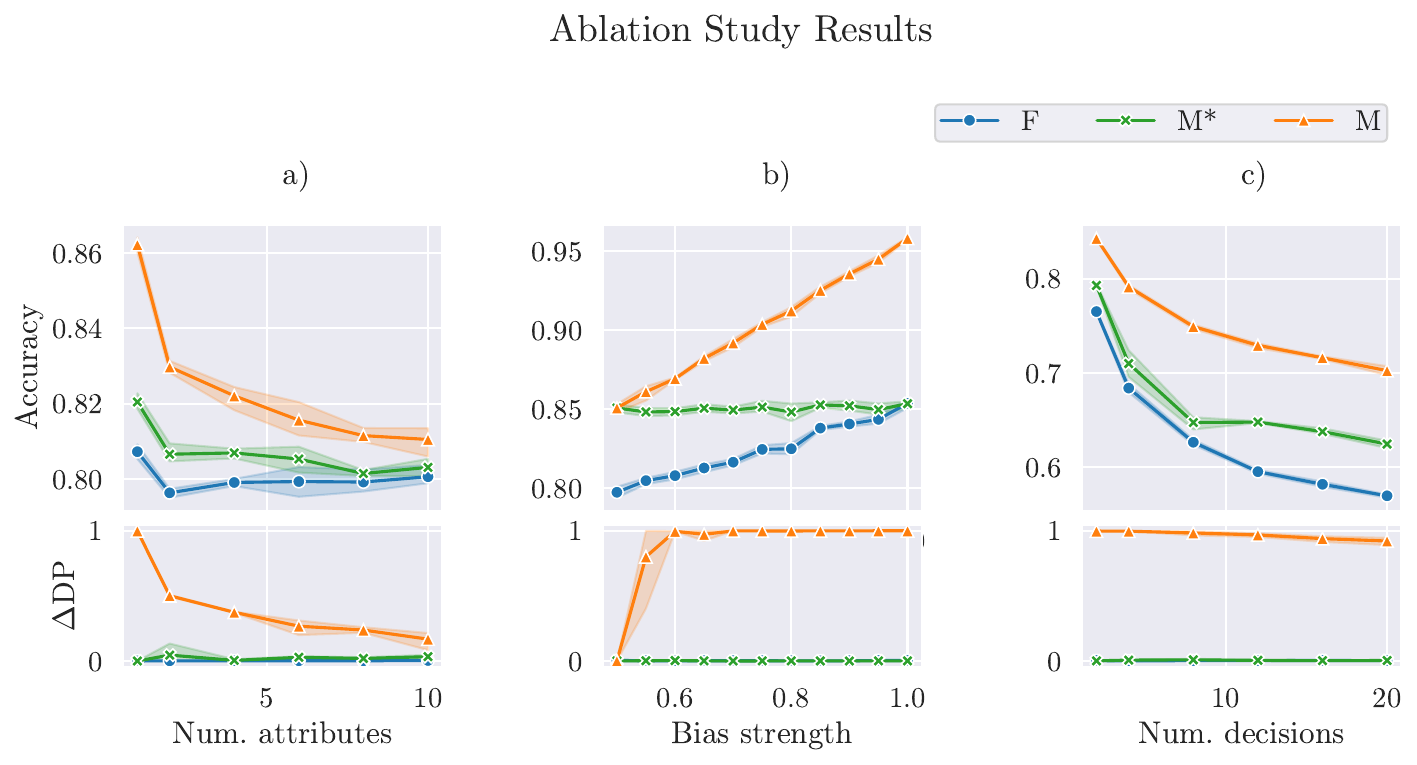}
    \caption{\small Results of our ablation studies. Accuracy of the modified prediction model $M^*$ is always better than using no sensitive attributes ($F$), while unfairness measured using the DP metric is minimized compared to the unfair model $M$.}
    \label{fig:results-all-ablations}

\end{figure}

\mypar{Ablation Studies}
We systematically investigated the effects of data characteristics on the results of our approach.
Fig.~\ref{fig:results-all-ablations} shows how accuracy and unfairness change when the number of sensitive attributes changes (Fig.~\ref{fig:results-all-ablations} \textit{a}), the strength of biases in decisions (Fig.~\ref{fig:results-all-ablations} \textit{b}), or the number of decisions using sensitive attributes (Fig.~\ref{fig:results-all-ablations} \textit{c}).
Each of these experiments uses 10,000 cases.

\mypartwo{Number of Sensitive Attributes}
Using the \textit{cs} process as a basis, we vary the number of sensitive attributes affecting the fairly and unfairly biased decision between 1 and 10.
We find that regardless of the number of sensitive attributes, model $M^*$ achieves higher accuracy than the baseline model $F$, while minimizing $\Delta \textit{DP}$ for all attributes.

\mypartwo{Bias Strength}
We vary the strength of the unfairly biased decisions in the \textit{cs} event log between 0.5 (perfectly equal) and 1.0 (perfectly biased).
As bias strength increases, the unfair model $M$ improves its accuracy, while that of model $M^*$ stays the same, which shows that our approach can remove unfair biases regardless of their strength.
This notion is further confirmed by minimal $\Delta{DP}$ for all levels of bias strength in the modified model $M^*$.
Interestingly, the accuracy of model $F$ also increases for stronger biases. 
This results from the structure of the underlying process, in which the unfairly biased decision directly allows making a prediction in the subsequent fairly biased decision.

\mypartwo{Number of Biased Decisions}
This experiment uses a synthetic event log of alternating fairly and unfairly biased decisions.
We vary the number of these decisions between 2 and 20, finding that our approach is again able to modify the unfair model $M$, such that unfairness is minimized, while outperforming the predictive performance of the fair baseline model $F$.
Note that the performance of all models degrades with an increasing number of decisions, as the event log complexity gradually exceeds the expressiveness of the neural network model we used.

\mypartwo{Effect of Removing Fine Tuning}  
Finally, we investigate the importance of fine tuning the original \textit{teacher} model using the \textit{student} model.
Using the distilled decision tree for predictions directly resulted in diminished performance, especially for complex event logs and event logs with weak biases.
This suggests that even a simple neural network model, such as the one we used in our experiments, is able to outperform decision tree models and learns patterns that are worth preserving by purposeful fine tuning.
We suspect that this effect is even more pronounced in more complex prediction models and for more complex event logs.

%
%
%
%%%%%%%%%%%%%%%%%%%%%%%%%%%%%%%%%%%%%%%%%%%%%
% CONCLUSION
%%%%%%%%%%%%%%%%%%%%%%%%%%%%%%%%%%%%%%%%%%%%%

\section{Concluding Remarks}
\label{sec:conclusion}
%\vspace{-0.2cm}
%
%
\mypar{Implications for Theory and Practice}
A first theoretical implication of our work is the recognition that not all biases in prediction models should be considered inherently negative. 
While biases are often viewed as a flaw that must be eliminated, certain biases can reflect domain-specific requirements that should be preserved. 
Our experiments have demonstrated that such a nuanced understanding of biases is beneficial, because removing only unwanted biases leads to more accurate and context-sensitive predictions. 

Even in scenarios where fairness-driven changes to the model are not strictly necessary, the insights gained from analyzing and addressing biases can play a critical role in building trust among stakeholders. Understanding how a black-box model behaves and ensuring it aligns with ethical and legal standards can foster confidence in its use, particularly in socially sensitive contexts such as hiring, healthcare, and law enforcement. 

Also with regard to legal and regulatory frameworks (e.g., \emph{European Union's Artificial Intelligence Act}\footnote{\url{https://artificialintelligenceact.eu/}, last accessed Mar. 18, 2025.}) that emphasize that \gls{ml} systems need to be transparent, unbiased, and equitable in their decision-making process, our approach provides a practical way to adjust prediction models to meet fairness standards, without compromising their accuracy unnecessarily.

%\vspace{-0.2cm}
\mypar{Limitations and Threats to Validity}
The effectiveness of our approach can be hampered because decision tree models may struggle to handle complex tasks with the same effectiveness as advanced machine learning models (e.g., deep learning models), leading to a problematic performance gap between the distilled decision tree model and the teacher model. However, our evaluation shows that for common real-life event logs, the capabilities of decision tree models are sufficient to approximate deep-learning prediction models. 

A potential second limitation arises with more activities, attributes, and dependencies, resulting in large, more intricate decision tree models that become difficult for domain experts to interpret and adjust. Therefore, it would be essential to provide appropriate tool support to handle the increasing complexity. This limitation is strongly tied to the fact that, in contrast to methods from related work, such as~\cite{qafari2019fairness,de2024achieving,peeperkorn2024}, our approach requires more direct input from domain experts. In practice, this means that our approach is likely to be more labor-intensive and consequently more cost-intensive than comparable methods. However, it comes with the advantage of being more precise and allows a more controlled removal of biases. Nevertheless, it should be noted that the method can theoretically also be used for the opposite, i.e., to willfully introduce biases into the model. 
A third limitation is that we focused exclusively on case attributes. 
However, event attributes could be encoded similarly to case attributes.

%\vspace{-0.2cm}
\mypar{Conclusion and Future Work}
In this paper, we proposed and evaluated a model-agnostic approach based on knowledge distillation that identifies biasing predictions in next-activity prediction models and offers a human-controlled option to rectify the models.
In contrast to previous works, this mechanism allows rectifying biases even in situations where a sensitive attribute is used both fairly and unfairly. 
Our evaluation shows that this approach efficiently captures biases in prediction models and performs better than discarding all sensitive attributes.
In future work, we plan to apply our methodology to additional prediction targets, such as outcome and remaining-time prediction, to assess its wider applicability.
We also intend to integrate event attributes to improve both fairness evaluation and model performance.
Since the method is not limited to neural networks, future research should investigate its effectiveness across various black-box models to determine how well it generalizes.
Finally, given that our approach depends strongly on expert manual inspection and domain knowledge, future work should include user studies to evaluate its practical usability and scalability in real-world settings.
%
%

%%%%%%%%%%%%%%%%%%%%%%%%%%%%%%%%%%%%%%%%%%%%%
% BIBLIOGRAPHY
%%%%%%%%%%%%%%%%%%%%%%%%%%%%%%%%%%%%%%%%%%%%%
\bibliographystyle{splncs04}
\bibliography{bibliography}
\end{document}